\documentclass[preprint,12pt,times]{elsarticle}

\usepackage{amssymb}
\usepackage{amsmath}

\usepackage{amsfonts}
\usepackage{algorithmic}
\usepackage{graphicx}
\usepackage{textcomp}
\usepackage{xcolor}

\usepackage{algorithm}

\usepackage{placeins}
\usepackage{multirow}
\usepackage{booktabs}       

\usepackage{nicefrac}       
\usepackage{microtype}      
\usepackage{epsfig}
\usepackage{subfigure}
\usepackage{newfloat}
\usepackage{listings}

\usepackage{natbib}

\journal{Neurocomputing}

\begin{document}

\begin{frontmatter}


\title{ACSAN: All-Convolutional Self-Attention Network Integrating Convolutional Operations with Vision Transformer}

\author{Zimian Wei\footnotemark[1], Tianqi Zhang\footnotemark[1], Hengyue Pan\footnotemark[2]}

\affiliation{organization={College of Computer Science and Technology, National University of Defense Technology},
	addressline={109 Deya Road}, 
	city={Changsha},
	postcode={410073}, 
	state={Hunan},
	country={China}}

\begin{abstract}
In the past few years, Vision Transformer (ViT) has achieved competitive performance in the field of computer vision. However, there still remain some problems. For instance, ViTs rely on large-scale training data since they simply contain a large number of trainable parameters, which slows down their learning efficiency and limits their applicability in resource-constrained scenarios. In this paper, we propose a novel deep learning model, namely All-Convolutional Self-Attention Network (ACSAN), which applies convolution operations to replace fully-connected matrices multiplications in the ViT framework. To implement ACSAN, we firstly design Dynamic Convolutional Self-Attention (DCSA) as the basic module, which employs multi-scale dilated convolutions to generate Query, Key, and Value. We define spatial self-attention and channel self-attention  to generate attention features respectively. The two attention features are finally merged as the output of the DCSA module. Based on DCSA, we construct the basic ACSAN block. Specifically, we use DCSA to replace the regular self-attention operations, and use convolution-based Equivalent Convolutional Fully Connected (EqConvFC) modules to replace MLPs. To evaluate the performance, we train ACSAN on CIFAR-10, CIFAR-100 and ImageNet-1k datasets from scratch. Experimental results show that ACSAN achieves good classification accuracy (the best variants achieve 93.52\% on CIFAR-10, 73.38\% on CIFAR-100, and 83.3\% on ImageNet-1k). Moreover, in comparison with standard ViT and other state-of-the-art methods, ACSAN uses fewer parameters to achieve comparable or better accuracy.

\end{abstract}

\begin{highlights}
\item We propose DCSA module that combines self-attention with convolution operation
\item We define a novel model, namely ACSAN, which is a fully convolutional self-attention network
\item ACSAN achieves good image classification accuracy over ViTs on several datasets
\end{highlights}


\end{frontmatter}

\footnotetext[1]{These authors contributed equally to this work.}
\footnotetext[2]{Corresponding author}



\section{Introduction}
\label{sec:introduction}

The past few years have witnessed the fast development of Transformer-related architectures. Starting from Transformer~\cite{vaswani2017attention}, many related models that apply the self-attention mechanism show outstanding performance in the field of natural language processing (NLP) due to their powerful global modeling ability from the self-attention mechanism~\cite{devlin2019bert,radford2018gpt,radford2019gpt2,brown2020gpt3}. Moreover, the introduction of Vision Transformer (ViT)~\cite{dosovitskiy2021vit} enables the Transformer models to deal with image-related tasks. To avoid large-scale matrices multiplication, ViT splits images into patch sequences and feeds them into a standard Transformer encoder for processing. A positional encoding mechanism is applied to keep the spatial information of the input images. In practice, ViT achieved state-of-the-art performance on image classification tasks. 

However, compared with CNNs~\cite{lecun1995convolutional}, the success of ViT highly relies on the large amount of training data, since the self-attentions and MLPs contain large amount of trainable parameters, and can hardly capture local features of input images. For instance, according to~\cite{dosovitskiy2021vit}, ViTs require large-scale pre-training to achieve good performance on ImageNet dataset. Moreover, due to the large number of matrices multiplication operations, the training and inference efficiency of ViTs is also not good enough. 

There are many works that try to introduce more local priors to provide ViTs with better ability to extract image features. For example, Swin Transformer~\cite{liu2021swin} builds hierarchical feature maps and a shifted window approach for better image classification abilities. \cite{lee2021vision} proposes Shifted Patch Tokenization (SPT) and Locality Self-Attention (LSA) to provide ViTs with locality inductive bias, which make ViTs achieve good performance when training data are not enough. In~\cite{yuan2021tokens}, the Tokens-to-Token Vision Transformer (T2T-ViT) is proposed. T2T-ViT recursively aggregates neighboring Tokens into one Token to model the local structure representation and reduce the length of tokens. In this way, T2T-ViT can achieve good image classification accuracy on ImageNet dataset when training from scratch.

The above-mentioned methods imply that providing local structure information for ViTs may help them to capture image features better. In fact, one main reason that CNN played an essential role in computer vision is that the image convolution operation has the ability to extract local image features. Therefore, some previous works consider incorporating image convolution operations into the ViT framework to find a better balance between the network performance and computation complexity. For instance, CvT~\cite{wu2021cvt} uses convolution operations to generate Q, K, and V for self-attention, while ACmix~\cite{pan2022acmix} applies convolution operations as another branch to work with self-attentions. UniFormer~\cite{li2023uniformer} designs a network module called Multi-Head Relation Aggregator (MHRA). MHRA unifies image convolution and self-attention, and achieves efficient and effective representation learning in image-related tasks. In~\cite{wei2023dmformer}, DMFormer is proposed. DMFormer introduces a novel dynamic multi-level attention (DMA) mechanism, which combines the advantages of convolution and self-attention. Based on DMA, DMFormer achieves good performance on image classification and segmentation tasks. 

However, even though there are many research progresses that can improve training efficiency and reduce data requirement of ViT, the challenges still remain. For instance, most of previous methods only apply convolutions as supplementary roles, while self-attentions are still implemented by using fully-connected structures, which means the number of trainable parameters remains large. Moreover, this makes the previous models can hardly deal with the cases that training data are extremely lacking (CIFAR-100 dataset for example, each class only contains around 500 low-resolution training samples).

Therefore, we hope to design a novel neural network architecture, which equips ViTs with better ability to extract local features of images. Different from previous works, we consider directly defining a self-attention computation on convolution operation. Specifically, we define channel self-attention and spatial self-attention on convolution operations, which extract local features and global dependencies simultaneously. Moreover, the MLP layers in ViTs, which contain large number of trainable parameters with high computational complexity, are also replaced by convolution-based Equivalent Convolutional Fully Connected (EqConvFC) modules. In this way, the proposed model has the ability to maintain strong representation capabilities while significantly reducing dependency on training data and improving computational efficiency. Generally speaking, the main contributions of this paper are as follows:

\begin{itemize}
	\item \textbf{DCSA Module}: We propose the Dynamic Convolutional Self-Attention (DCSA) module, which integrates convolutional operations into the self-attention mechanism, which provides both the strong local inductive bias of convolutional neural networks and the strong long-range dependency modeling capability of self-attention mechanisms. Moreover, DCSA introduces multi-scale dilated convolutions with the capability to capture multi-scale features of input images.
	
	\item \textbf{ACSAN Network}: Based on the proposed DCSA module, we construct the All-Convolutional Self-Attention Network (ACSAN), which combines the advantages of CNNs and ViTs, since it can simultaneously extract local and global features of input images.
	
	\item \textbf{Model Performance}: We perform experiments on several widely used image classification datasets, which include CIFAR-10, CIFAR-100, and ImageNet-1k. The experimental results imply that ACSAN can effectively handle scenarios with insufficient training samples. Without large-scale pre-training, ACSAN can achieve good classification performance with higher training efficiency.
\end{itemize}

The remainder of this paper is organized as follows. Section~\ref{sec:relwork} reviews the related works of this research. Section~\ref{sec:method} presents the details of the proposed DCSA module and ACSAN architecture. Section~\ref{sec:experiments} presents experimental results and ablation studies. Section~\ref{sec:conclusion} concludes the paper.

\section{Related Works}
\label{sec:relwork}

\subsection{Convolutional Neural Network}

Convolutional Neural Network (CNN)~\cite{lecun1995convolutional} is the most widely-used neural network architecture in the field of computer vision~\cite{girshick2015fast,niu2021image,pan2022fixed}. Inspired by biological visual systems, its local receptive field, parameter sharing mechanism, and hierarchical feature extraction make it suitable to extract image local features and reduce the computation and storage costs. The past few decades have witnessed the success of many CNN-based models, such as AlexNet~\cite{krizhevsky2012imagenet}, VGG~\cite{simonyan2014vgg}, ResNet~\cite{he2016resnet}, and ConvNext~\cite{liu2022convnet}. Even though currently self-attention based models (such as Vision Transformer) and their variants show competitive performance, CNNs are still the main backbone architecture of most image-related applications. Therefore, it is worthwhile to consider how to introduce the advantages of CNNs to the self-attention mechanism for better performance and efficiency.

\subsection{Vision Transformer}

Vision Transformer (ViT)~\cite{dosovitskiy2021vit} successfully transfers the Transformer architecture from natural language processing tasks to the field of computer vision. The core idea of ViT is to treat images as sequence data and use the self-attention mechanism to directly model global dependencies between image patches. In scenarios with sufficient training data, or when large-scale pre-trained models are available, ViT-based models achieve better performance over convolutional neural networks in many image-related tasks. However, when training data is limited, due to the lack of the strong inductive bias on image spatial structure, the performance of ViT models may suffer from noticeable drop. Another significant issue of ViT is the high training cost because of its large number of trainable parameters, which makes them hard to be trained and deployed on platforms with limited computational resources. Therefore, there are many previous works~\cite{liu2021swin,lee2021vision,yuan2021tokens,he2026linear} that try to reduce the data requirement of ViTs and achieve good results. 

There are also some works that begin to combine image convolution operations with attention mechanism, to make the model capture both local features and global dependencies. Early research such as BAM~\cite{park2018bam} and CBAM~\cite{woo2018cbam} insert channel and spatial attention modules into CNNs. SENet~\cite{hu2020senet} and ECA-Net~\cite{wang2020ecanet} applied channel attention branches into CNNs to enhance network performance. The combination of convolution operations and self-attentions has also been viewed as a worthwhile research topic. In~\cite{bello2019attention}, an attention-augmented convolution module is proposed, which contains two branches: one regular image convolution branch and one self-attention branch. The convolution feature and self-attention feature are finally concatenated as the output feature. Moreover, several studies further consider unifying convolution operations and self-attentions. CvT~\cite{wu2021cvt} introduces Convolutional Token Embedding and Convolutional Projection into self-attention mechanism. Specifically, the input feature maps are firstly processed by convolution layers, thus the token feature dimensions and number of channels can be adjusted. Then, the embedded features are projected to Q, K, and V by using depthwise convolutions. After that, the regular self-attention is calculated by using the generated Q, K, and V. In~\cite{pan2022acmix}, it is proved that one convolution layer can be split into a group of 1 $\times$ 1 convolution operations followed by convolution aggregation, while one self-attention module can also be decomposed into 1 $\times$ 1 convolution operations and self-attention aggregation. Based on this observation, ACMix is proposed, which firstly performs a group of 1 $\times$ 1 convolutions to extract image features, then introduces one convolution aggregation branch and one self-attention aggregation branch. In this way, image convolution and self-attention can be combined into one module. In~\cite{li2023uniformer}, UniFormer is proposed to unify the merits of convolution and self-attention in the ViT framework. Specifically, the core component of UniFormer, i.e., Multi-Head Relation Aggregator (MHRA), contains two variants. Local MHRA can be explained as an extension of MobileNet Block, which has the ability to capture local dependencies. While global MHRA shares features with self-attention, which enables it to extract global information. DMFormer~\cite{wei2023dmformer} contains Dynamic Multi-level Attention (DMA) modules, where multi-scale dilated convolutions are used to capture different scales of image local features, while a gating mechanism with fully-connected layers is introduced to generate attention information. The works above show the potential of the combination of image convolution and self-attention, which can be viewed as an important basis of our research.

\section{Methodology}
\label{sec:method}

In this section, we firstly propose the basic idea of DCSA, then present the design of ACSAN.

\subsection{DCSA Module}

\begin{figure}[t]
	\centering
	\includegraphics[width=1\linewidth]{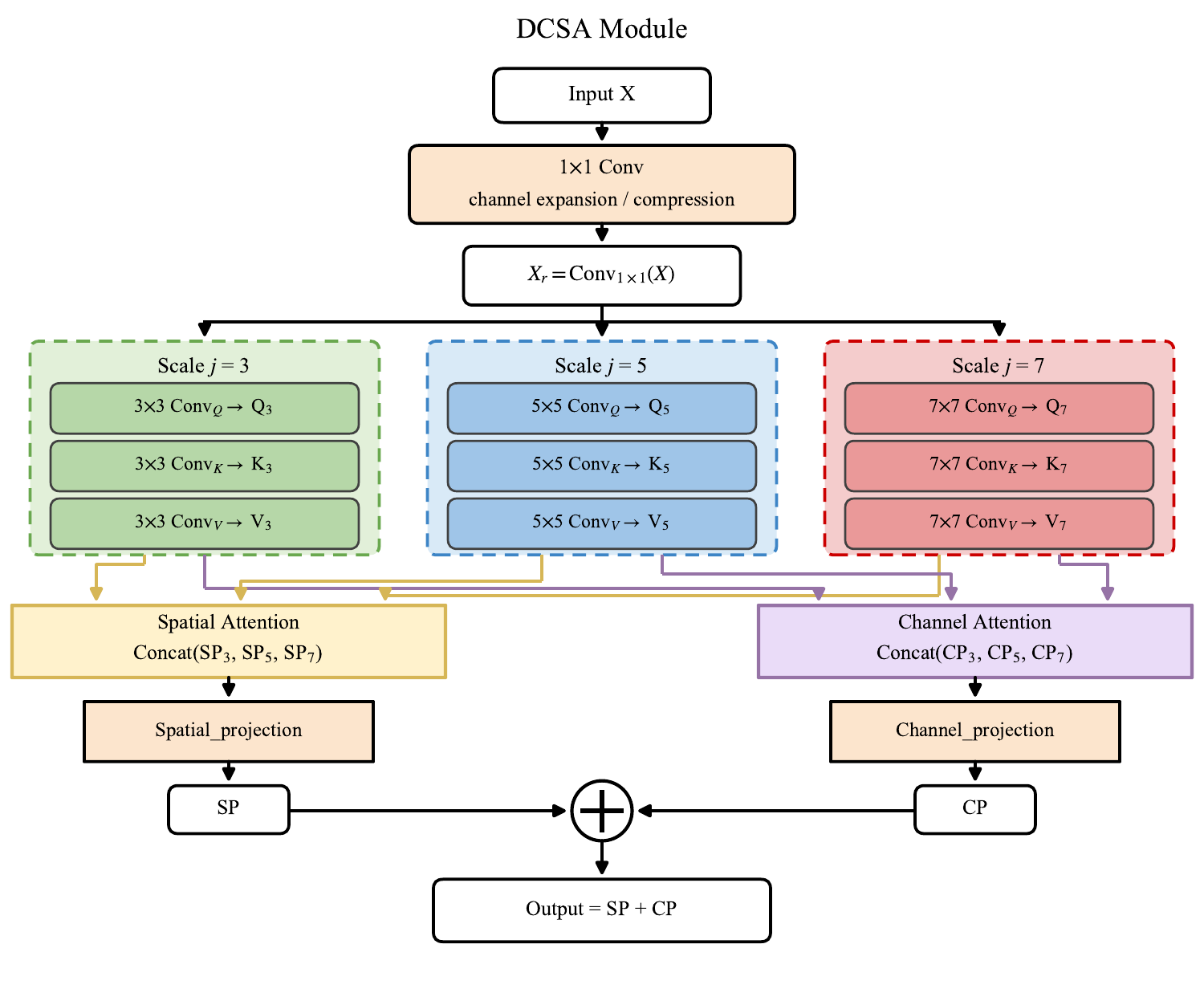}
	\caption{Structure of the Dynamic Convolutional Self-Attention (DCSA) module. The DCSA module generates Q, K, V via multi-scale dilated convolutions, then applies parallel spatial and channel attention branches before fusion.}
	\label{fig:dcsa}
\end{figure}

DCSA is the fundamental computational module of the proposed ACSAN model, which combines the self-attention mechanism with image convolution operation to endow the model with the ability to capture both local and global features. DCSA includes the following key components: (1) multi-scale dilated convolution unit (MDCU), (2) convolution self-attentions, and (3) feature fusion unit. The overall structure of DCSA is shown in Fig.~\ref{fig:dcsa}. 

\subsubsection{Multi-scale Dilated Convolution Unit}

The main objective of the proposed MDCU is to generate Query, Key, and Value for the further self-attention calculation. Compared with the regular self-attention mechanism, we use multi-scale dilated convolutions as a generator, instead of fully-connect matrices multiplication, for better ability to use local features of input feature maps. In one MDCU, the input feature maps are firstly processed by a 1 $\times$ 1 convolution for channel expansion/compression. Assuming that the input of MDCU is $X$, then we have:

\begin{equation}
	X_{r} = Conv_{1 \times 1}(X)
	\label{eq:MDCU1}
\end{equation}

Assuming that $X$ contains $c_i$ feature maps, while $X_{r}$ contains $c_o$ feature maps, then we have $c_o = r * c_i$. Here $r$ is a pre-defined expansion/compression rate. 

After that, inspired by DMFormer~\cite{wei2023dmformer}, we apply multi-scale dilated convolutions (we adopt 3, 5, and 7 in this paper) to generate a group of Query\-Key\-Value triplets:

\begin{equation}
	\label{eq:MDCU2}
	\begin{split}
		& Q_j = W_{j}^q * X_{r}  \\
		& K_j = W_{j}^k * X_{r}  \\
		& V_j = W_{j}^v * X_{r}
	\end{split}
\end{equation}

where $ W_{j}^*$ is the convolution kernel to generate $Q_j$, $K_j$ and $V_j$, and $j$ denotes the convolution kernel size of the dilated convolution.

\subsubsection{Convolution Self-Attentions}

For each group of $Q_j$, $K_j$ and $V_j$, we define two convolution self-attention branches: 

\textbf{Channel Self-Attention Branch: } We define the channel self-attention branch to capture the global dependency between each channel of the input feature maps. Assuming that the feature maps $Q_j$, $K_j$ and $V_j$ have the shape $(B, C, H, W)$, where $B$ is the batch size, $C$ is the number of channels, $H$ is the height, and $W$ is the width. Based on the definition of self-attention, we firstly flatten them to the shape $(B, C, H \times W)$:

\begin{equation}
	\label{eq:channelattn1}
	Q_f = \text{Flatten}(Q_j), \quad K_f = \text{Flatten}(K_j), \quad V_f = \text{Flatten}(V_j)
\end{equation}

In the flattened feature maps, the last dimension corresponds to all elements from the same channel. To calculate the channel self-attention, we transpose the last two dimensions of $K_f$ and obtain $K_f^T$ with shape $(B, H \times W, C)$, and then we have:

\begin{equation}
	\label{eq:channelattn2}
	\text{CAttn}(Q, K, V) = \text{Reshape}\left(\text{softmax}\left(\frac{Q_f K_f^T}{\sqrt{C}}\right) V_f\right)
\end{equation}

In Eq.~\ref{eq:channelattn2}, computing $Q_f \cdot K_f^T$ yields a $C-by-C$ matrix that semantically represents the co-dependency between channels. After scaling by $\sqrt{C}$ and softmax normalization, the channel attention weight matrix is obtained. Finally, we multiply the channel attention weight matrix by $V_f$, and reshape back to the size of $(B, C, H, W)$.

\textbf{Spatial Self-Attention Branch: } We define the spatial self-attention branch to capture the global dependency between each spatial location of the input feature maps. Similar to channel self-attention, we firstly get $Q_f$, $K_f$, and $V_f$ using Eq.~\ref{eq:channelattn1}. Different from Eq.~\ref{eq:channelattn2}, the calculation of spatial attention follows Eq.~\ref{eq:spatialattn}:

\begin{equation}
	\label{eq:spatialattn}
	\text{SAttn}(Q, K, V) = \text{Reshape}\left(\text{softmax}\left(\frac{Q_f^T K_f}{\sqrt{H \times W}}\right) V_f\right)
\end{equation}

where $Q_f^T$ has the shape of $(B, H \times W, C)$. Therefore, computing $Q_f^T \cdot K_f$ yields a $(H \times W)-by-(H \times W)$ matrix that semantically represents the co-dependency between spatial locations. After scaling by $\sqrt{H \times W}$ and softmax normalization, the spatial attention weight matrix is obtained. Finally, the spatial attention weight matrix should also be multiplied by $V_f$, and reshape back to the size of $(B, C, H, W)$.

\subsubsection{Feature Fusion Unit}

At the last stage of the DCSA module, we need to merge the results of channel self-attention and spatial self-attention. Firstly, we introduce two convolution projection layers, which include convolution operation, batch normalization, and ReLU, to project ${CAttn}(Q, K, V)$ and ${SAttn}(Q, K, V)$ to a pre-defined number of channels:

\begin{equation}
	\label{eq:FFU1}
	\begin{split}
		& CP = \text{Proj}(\text{CAttn}(Q, K, V))  \\
		& SP = \text{Proj}(\text{SAttn}(Q, K, V))  
	\end{split}
\end{equation}

Then, $CP$ and $SP$ are summed to obtain the final output of the DCSA unit:

\begin{equation}
	\label{eq:FFU2}
	\text{DCSA}(X) = CP + SP
\end{equation}

Based on the definition of DCSA module, we can learn that the introduction of self-attention enables the convolution operation to capture global features of input feature maps through two different directions (channel-wise and spatial-wise). This ability may help the whole ACSAN model to achieve better classification performance.

\subsubsection{Multi-Head DCSA}

Following the basic idea of ViT and related studies, DCSA can further enhance feature representation capability by introducing the multi-head self-attention mechanism. To implement multi-head convolutional self-attention without significantly increasing model parameters, we split the input feature $X$ by channels into $n$ sub-features (i.e., $n$ self-attention heads), then independently execute multi-scale convolution operations, self-attention computation, and projection. In this way, we get $n$ $CP$s and $n$ $SP$s. All $CP$s and $SP$s are then concatenated along the channel dimension:

\begin{equation}
	CP = \text{cat}({CP}_1, {CP}_2, \ldots, {CP}_n)
\end{equation}

\begin{equation}
	SP = \text{cat}({SP}_1, {SP}_2, \ldots, {SP}_n)
\end{equation}

Notice that inside each DCSA module, all convolution operations do not change the sizes of feature maps, thus the above-mentioned feature concatenation can be performed without any pre-processing. After the feature concatenation, we use Eq.~\ref{eq:FFU2} to get the final output of the multi-head DCSA. 

\subsection{ACSAN Network}

\begin{figure}[tbp]
	\centering
	\includegraphics[width=0.8\linewidth]{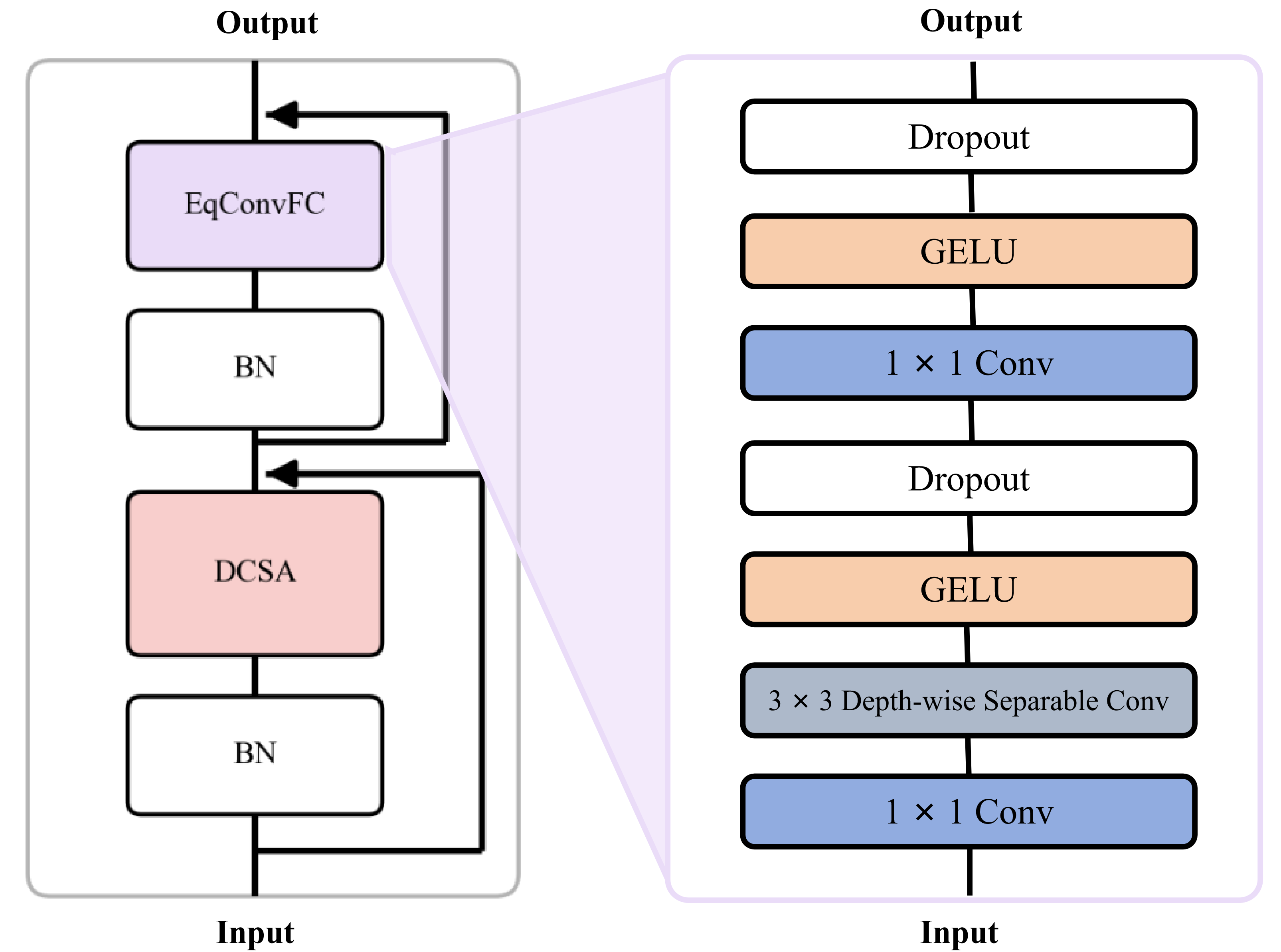}
	\caption{Left: the structure of a DCSA block, which contains a DCSA module and an EqConvFC module, each followed by batch normalization, with residual connections. Right: the detailed structure of an EqConvFC module. }
	\label{fig:dcsa_block}
\end{figure}

Based on the definition of DCSA, we design DCSA Block as the basic component of the ACSAN model. Each DCSA Block includes a DCSA module, an Equivalent Convolutional Fully Connected (EqConvFC) module, and batch normalization layers. Residual connections are adopted to alleviate the gradient vanishing problem  (see the left side of Fig.~\ref{fig:dcsa_block} for more details). Here the MLP layers in ViTs are replaced by EqConvFC modules, and each EqConvFC contains two regular $1\times1$ convolution layers, one $3\times3$ depth-wise separable convolution layer, activation functions (here we use GELU), and dropout layers (here we set dropout rate = 0.2) (see the right side of Fig.~\ref{fig:dcsa_block} for more details).

\begin{figure}[t]
	\centering
	\includegraphics[width=1\linewidth]{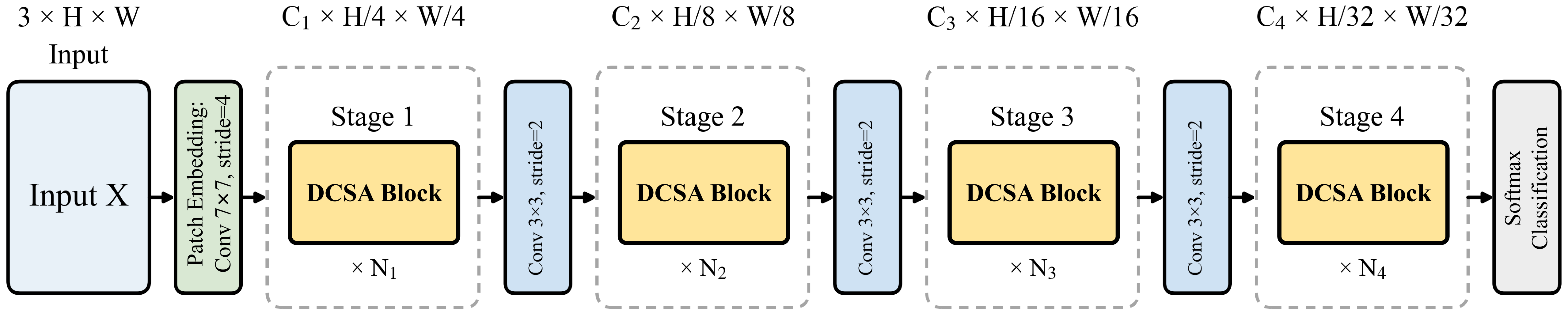}
	\caption{Overall architecture of the proposed ACSAN model. The model consists of multiple stages of DCSA Blocks with downsampling between stages.}
	\label{fig:overview}
\end{figure}

The proposed ACSAN architecture uses DCSA Blocks as basic modules. The input images are firstly embedded by using a $7\times7$ convolution layer with $stride=4$, and then propagated through several network stages. Within each network stage, one or more DCSA Blocks are directly concatenated with feature map sizes unchanged. Between different stages, 3$\times$3 convolutions with stride 2 are used for downsampling and channel number changing. In the classification layer, a fully-connected layer with softmax is used to convert DCSA module outputs to class feature vectors. The overall structure of ACSAN is shown in Fig.~\ref{fig:overview}.

To evaluate the practical performance of models with different number of parameters, we design three scales of ACSAN models for CIFAR-10 and CIFAR-100 datasets, i.e., ACSAN\_small, ACSAN\_base, and ACSAN\_large. The channel expansion/compression ratio $r$ of DCSA modules is set to 0.75 to reduce model parameters while effectively improving training and inference efficiency. The number of heads in the multi-head attention is set to 4. Detailed architecture configurations are shown in Table~\ref{tab:acsan_arch}.

\begin{table}[t]
	\centering
	\caption{Detailed architecture configurations of ACSAN variants for CIFAR-10 and CIFAR-100.}
	\label{tab:acsan_arch}
	\begin{tabular}{l|c|c|c}
		\toprule
		\textbf{Stage} & \textbf{ACSAN\_small} & \textbf{ACSAN\_base} & \textbf{ACSAN\_large}  \\
		\midrule
		Stage 1 & [Block]$\times$2, $C$=64 & [Block]$\times$2, $C$=64  & [Block]$\times$2, $C$=64 \\
		Stage 2 & [Block]$\times$2, $C$=128 & [Block]$\times$2, $C$=128 & [Block]$\times$2, $C$=128  \\
		Stage 3 & [Block]$\times$3, $C$=256 & [Block]$\times$4, $C$=256 & [Block]$\times$6, $C$=256  \\
		Stage 4 & [Block]$\times$2, $C$=320 & [Block]$\times$2, $C$=352 & [Block]$\times$2, $C$=320 \\
		\bottomrule
	\end{tabular}
\end{table}

For ImageNet, we design two ACSAN models, i.e., ACSAN\_ImageNet\_base and ACSAN\_ImageNet\_large. Specifically, ACSAN\_ImageNet\_base follows ACSAN\_large settings, while ACSAN\_ImageNet\_large extends per-stage channel counts to 64, 128, 320, 512 for richer feature extraction. ViT\_ImageNet\_base has 10 layers, 6 heads, hidden dimension 384, and MLP ratio 2. ViT\_ImageNet\_large has 12 layers, 6 heads, hidden dimension 384, and MLP ratio 4. Both models use patch size 8 for fine-grained classification. Detailed architecture configurations are shown in Table~\ref{tab:acsan_arch2}.

\begin{table}[t]
	\centering
	\caption{Detailed architecture configurations of ACSAN variants for ImageNet dataset.}
	\label{tab:acsan_arch2}
	\begin{tabular}{l|c|c}
		\toprule
		\textbf{Stage} &  \textbf{ACSAN\_ImageNet\_base} & \textbf{ACSAN\_ImageNet\_large} \\
		\midrule
		Stage 1 & [Block]$\times$2, $C$=64 & [Block]$\times$2, $C$=64\\
		Stage 2 & [Block]$\times$2, $C$=128 &  [Block]$\times$2, $C$=128 \\
		Stage 3 & [Block]$\times$6, $C$=256 & [Block]$\times$6, $C$=320 \\
		Stage 4 & [Block]$\times$2, $C$=320 &  [Block]$\times$2, $C$=512\\
		\bottomrule
	\end{tabular}
\end{table}

Notice that during the training process, all models are randomly initialized without pre-training.

\section{Experimental Results}
\label{sec:experiments}

\subsection{Datasets and Implementation Details}

In this paper, we apply three widely used datasets to evaluate the proposed ACSAN, i.e., CIFAR-10~\cite{krizhevsky2009cifar}, CIFAR-100~\cite{krizhevsky2009cifar} and ImageNet~\cite{krizhevsky2012imagenet}. 

The proposed ACSAN network is implemented using PyTorch 2.80~\cite{paszke2019pytorch} in a Python 3.10 environment. The computation platform includes Intel Xeon 4108 CPU, 256 GB memory, and eight Tesla A100 GPUs. During the training process of all datasets, pre-trained models are not applied for ACSAN and all baseline models. 

\begin{table}[h]
	\centering
	\caption{Comparison of ACSAN and ViT on CIFAR-10 dataset. All models are trained from scratch without pre-training.}
	\label{tab:cifar10}
	\begin{tabular}{l|c|c|c}
		\toprule
		\textbf{Model} & \textbf{Params (M)} & \textbf{FLOPs (M)} & \textbf{Top-1 Acc. (\%)} \\
		\midrule
		ViT\_small    & 8.04  & 541.31 & 90.61 \\
		ViT\_base     & 9.82  & 661.34 & 90.50 \\
		ViT\_large    & 10.71 & 721.36 & 90.63 \\
		\midrule
		ACSAN\_small  & 8.03  & 53.84  & 93.26 \\
		ACSAN\_base   & 9.75  & 58.87  & 93.52 \\
		ACSAN\_large  & 11.06 & 66.68  & 93.38 \\
		\bottomrule
	\end{tabular}
\end{table}

\subsection{CIFAR-10 Experiments}

In our CIFAR-10 experiments, all models are trained using the AdamW optimizer for 250 epochs with a batch size of 64 on a single GPU. The initial learning rate was set to $1 \times 10^{-3}$ with cosine annealing scheduling. The baseline ViT models are configured following~\cite{dosovitskiy2021vit} with adjustments for CIFAR datasets, and we make them have a similar number of trainable parameters with the ACSAN counterparts for fair performance comparison. Specifically, ViT\_small has 9 layers with 12-head self-attention; ViT\_base has 11 layers with 12-head self-attention; ViT\_large has 12 layers with 12-head self-attention. All three ViT models have a hidden dimension of 384 and MLP ratio of 1 (the main objective of these configurations is to guarantee that each regular ViT has a similar number of parameters as the corresponding ACSAN model), with patch size of 4. All ACSAN and ViT models are randomly initialized without any pre-training, and share the same training configuration. Table~\ref{tab:cifar10} summarizes the CIFAR-10 classification results. 

The experimental results reflect two important facts: 
\begin{itemize}
	\item In comparison with the ViT model with a similar number of parameters, ACSAN can obviously reduce the computation complexity, which is meaningful for real-world applications. 
	\item Due to the fully-convolution structure, all ACSANs achieve better performance compared with their ViT counterparts. 
\end{itemize}

\begin{table}[t]
	\centering
	\caption{Comparison of ACSAN and ViT on CIFAR-100 dataset. All models are trained from scratch without pre-training.}
	\label{tab:cifar100}
	\begin{tabular}{l|c|c|c}
		\toprule
		\textbf{Model} & \textbf{Params (M)} & \textbf{FLOPs (M)} & \textbf{Top-1 Acc. (\%)} \\
		\midrule
		ViT\_small  & 8.07  & 541.35 & 65.50 \\
		ViT\_base   & 9.85  & 661.38 & 65.02 \\
		ViT\_large  & 10.74 & 721.39 & 65.15 \\
		\midrule
		ACSAN\_small  & 8.05  & 53.87 & 73.38 \\
		ACSAN\_base   & 9.78  & 58.90 & 72.91 \\
		ACSAN\_large  & 11.09 & 66.71 & 73.36 \\
		\bottomrule
	\end{tabular}
\end{table}

\subsection{CIFAR-100 Experiments}

Our CIFAR-100 experiments share the same training settings as CIFAR-10 experiments. Table~\ref{tab:cifar100} summarizes the CIFAR-100 results. From the experimental results we can learn that the proposed ACSAN model can also handle the cases that the training sample size of each class is extremely small without any pre-training. From the experimental results of CIFAR-100, we can learn that under the condition of the extreme scarcity of training samples, the proposed ACSAN models substantially improve the classification accuracy and reduce the computation complexity in comparison with the corresponding regular ViT models with a similar number of model parameters. The good results can also be attributed to the combination of self-attention and image convolution.

\begin{table}[t]
	\centering
	\caption{Comparison of ACSAN and baseline models on ImageNet dataset. All ViTs and ACSANs are trained from scratch.}
	\label{tab:imagenet}
	\begin{tabular}{l|c|c|c}
		\toprule
		\textbf{Model} & \textbf{Params (M)} & \textbf{FLOPs (G)} & \textbf{Top-1 Acc. (\%)} \\
		\midrule
		ViT\_ImageNet\_base               & 12.24 & 14.05 &   81.1   \\
		CvT-13~\cite{wu2021cvt}           & 20.0  & 4.5   &  81.6\\
		T2T-ViT-14~\cite{yuan2021tokens}  & 21.5  & 4.8   &  81.5\\
		Swin-T~\cite{liu2021swin}         & 28.3  & 4.5   &  81.3\\
		Swin-T-ACMix~\cite{pan2022acmix}  & 30.0  & 4.6   &  81.9\\
		ConvNext-T~\cite{liu2022convnet}   & 28.6  & 4.5   &  82.1\\
		UniFormer~\cite{li2023uniformer}  & 21.5  & -   &  {\bf 82.9}\\
		DMFormer-S~\cite{wei2023dmformer} & 26.7  & 5.0   &  82.8\\
		WeightFormer-S~\cite{he2026linear} & 27.0  & 4.4  &  81.3\\
		\midrule
		ACSAN\_ImageNet\_base            & {\bf 11.08} & 9.77   &   82.7 \\
		\midrule	
		ViT\_ImageNet\_large             & 21.69 & 22.40 &    81.9  \\	
		CvT-21~\cite{wu2021cvt}          & 32.0  & 7.1   &  82.5\\	
		T2T-ViT-24~\cite{yuan2021tokens}  & 64.1  & 13.8   &  82.3\\	
		Swin-S~\cite{liu2021swin}        & 49.6  & 8.7   &  83.0\\	
		Swin-S-ACMix~\cite{pan2022acmix}  & 51.0  & 9.0   &  83.5\\	
		ConvNext-S~\cite{liu2022convnet}   & 50.0  & 8.7   &  83.1\\	
		DMFormer-L~\cite{wei2023dmformer} & 45.0  & 8.7   &  {\bf 83.6}\\
		\midrule
		ACSAN\_ImageNet\_large            & {\bf 19.94} & 10.82 &   83.3\\
		\bottomrule
	\end{tabular}
\end{table}

\subsection{ImageNet Experiments}

In our ImageNet experiments, we use the AdamW optimizer to train 300 epochs on 8 Tesla A100 GPUs from scratch, and the training batch size is set to 64. The initial learning rate is $2 \times 10^{-4}$ with a cosine annealing schedule.  Here, besides the corresponding ViT structure, we also apply several previous studies as baselines to compare with the proposed ACSAN.

Table~\ref{tab:imagenet} shows the experimental results, and from which we can learn that the proposed ACSAN models show better classification accuracy and lower computation FLOPs compared with corresponding ViT structure with a similar number of model parameters. This fact implies that the proposed method also has potential to deal with large-scale image classification tasks. Moreover, compared with other third-party baselines, ACSAN also shows its advantages to achieve good balance between computation/storage costs and network performance (For WeightFormer~\cite{he2026linear}, we only select WeightFormer-S as one of the baselines, since WeightFormer-L applies a higher resolution of input images (448-by-448 instead of 224-by-224), which makes it impossible to compare directly with ACSANs and other baselines). For example, compared with Swin Transformer~\cite{liu2021swin}, CvT~\cite{wu2021cvt}, T2T-ViT~\cite{yuan2021tokens}, ConvNext~\cite{liu2022convnet}, UniFormer~\cite{li2023uniformer}, and WeightFormer~\cite{he2026linear}, the proposed ACSAN models achieve better classification performance with lower number of model parameters at moderate extra FLOPs. In comparison with ACMix~\cite{pan2022acmix} and DMFormer~\cite{wei2023dmformer}, we can learn that ACSAN\_base shows comparable performance with lower number of model parameters.

\begin{table}[t]
	\centering
	\caption{Ablation study results on CIFAR-10 dataset.}
	\label{tab:ablation_dcsa}
	\begin{tabular}{l|c|c}
		\toprule
		\textbf{Model} & \textbf{Params (M)} & \textbf{CIFAR-10 (\%)}  \\
		\midrule
		Channel Attn Only            & 9.78 & 93.28  \\
		Spatial Attn Only            & 9.78 & 93.29  \\
		ACSAN\_base                  & 9.78 & 93.52  \\
		\midrule
		Fully Convolution            & 9.58 & 89.47  \\
		Fully Depthwise Sep. Conv    & 9.52 & 89.14  \\
		DMFormer                     & 9.85 & 88.70  \\
		\bottomrule
	\end{tabular}
\end{table}

\subsection{Ablation Studies}

To prove the advantages brought by the main innovations of this research, we conduct ablation studies. In our ablation studies, we use ACSAN\_base as the baseline method, and design several ablation models. Firstly, to verify the impact of the proposed spatial and channel self-attention, we introduce two extra models:

\begin{itemize}
	\item 'Channel Attn Only' model: the two self-attention branches in each DCSA module are all channel self-attentions
	\item 'Spatial Attn Only' model: the two self-attention branches in each DCSA module are all spatial self-attentions
\end{itemize} 

In this way, we can show the impacts of the combination of two attentions. 

Moreover, we also consider three ablation models that all self-attentions are removed:

\begin{itemize}
	\item 'Fully Convolution' model: all self-attention computations in the ACSAN\_base are removed.  
	\item 'Fully Depthwise Sep. Conv' model: in comparison with the 'Fully Convolution' model, the dilated convolutions in each Multi-scale Dilated Convolution Unit are replaced by depthwise separable convolutions. 
	\item DMFormer~\cite{wei2023dmformer}: the only difference between DMFormer and 'Fully Depthwise Sep. Conv' model is that all MLP layers in DMFormer are replaced by convolution layers in the 'Fully Depthwise Sep. Conv' model. 
\end{itemize} 

Table~\ref{tab:ablation_dcsa} presents the experimental results of ablation studies on CIFAR-10 dataset. Considering the results of 'Channel Attn Only' model, 'Spatial Attn Only' model and ACSAN\_base, we can learn that the combination of channel self-attention and spatial self-attention brings about positive impacts on the network performances. Moreover, the results of 'Fully Convolution' model, 'Fully Depthwise Sep. Conv' model, and DMFormer show that the proposed DCSA module as well as the fully-convolution ACSAN network can substantially improve the image classification performance when training data is not enough.

\section{Conclusion and Future Works}
\label{sec:conclusion}

This paper has proposed ACSAN, which fuses ViT's core self-attention mechanism with convolution operations to form convolutional self-attention. Based on this, we have proposed the DCSA module, which contains multi-scale dilated convolution unit (MDCU), convolution self-attentions, and feature fusion unit. By using DCSA modules and equivalently replacing ViT's MLP layers with convolution layers, we form the All-Convolutional Self-Attention Network (ACSAN). Experimental results show that ACSAN achieves strong performance compared with baseline models under the condition that the training data are not enough. The potential future works of this work include generalizing the proposed method to more application fields such as object detection and tracking, and applying the method to large language models.

\bibliographystyle{elsarticle-num}
\bibliography{references}

@article{lee2021vision,
	title={Vision transformer for small-size datasets},
	author={Lee, Seung Hoon and Lee, Seunghyun and Song, Byung Cheol},
	journal={arXiv preprint arXiv:2112.13492},
	year={2021}
}

@inproceedings{wu2021cvt,
	title={Cvt: Introducing convolutions to vision transformers},
	author={Wu, Haiping and Xiao, Bin and Codella, Noel and Liu, Mengchen and Dai, Xiyang and Yuan, Lu and Zhang, Lei},
	booktitle={2021 IEEE/CVF International Conference on Computer Vision (ICCV)},
	pages={22--31},
	year={2021},
	organization={IEEE}
}

@article{niu2021image,
	title={Image translation between high-resolution optical and synthetic aperture radar (SAR) data},
	author={Niu, Xin and Yang, Di and Yang, Ke and Pan, Hengyue and Dou, Yong and Xia, Fei},
	journal={International Journal of Remote Sensing},
	volume={42},
	number={12},
	pages={4758--4784},
	year={2021},
	publisher={Taylor \& Francis}
}

@article{pan2022fixed,
	title={Fixed-Size Objects Encoding for Visual Relationship Detection},
	author={Pan, Hengyue and Niu, Xin and Shen, Siqi and Chen, Yixin and Qiao, Peng and Huang, Zhen and Li, Dongsheng},
	journal={Neural Processing Letters},
	volume={54},
	number={4},
	pages={3249--3261},
	year={2022},
	publisher={Springer}
}

@inproceedings{girshick2015fast,
	title={Fast r-cnn},
	author={Girshick, Ross},
	booktitle={Proceedings of the IEEE international conference on computer vision},
	pages={1440--1448},
	year={2015}
}

@article{he2026linear,
	title={Linear-Time Global Visual Modeling without Explicit Attention},
	author={He, Ruize and Han, Dongchen and Huang, Gao},
	journal={arXiv preprint arXiv:2605.01711},
	year={2026}
}

@inproceedings{liu2022convnet,
	author={Liu, Zhuang and Mao, Hanzi and Wu, Chao-Yuan and Feichtenhofer, Christoph and Darrell, Trevor and Xie, Saining},
	booktitle={2022 IEEE/CVF Conference on Computer Vision and Pattern Recognition (CVPR)}, 
	title={A ConvNet for the 2020s}, 
	year={2022},
	volume={},
	number={},
	pages={11966-11976},
	doi={10.1109/CVPR52688.2022.01167}}

@inproceedings{yuan2021tokens,
	title={Tokens-to-token vit: Training vision transformers from scratch on imagenet},
	author={Yuan, Li and Chen, Yunpeng and Wang, Tao and Yu, Weihao and Shi, Yujun and Jiang, Zi-Hang and Tay, Francis EH and Feng, Jiashi and Yan, Shuicheng},
	booktitle={Proceedings of the IEEE/CVF international conference on computer vision},
	pages={558--567},
	year={2021}
}

@article{lecun1995convolutional,
	title={Convolutional networks for images, speech, and time series},
	author={LeCun, Yann and Bengio, Yoshua},
	journal={The handbook of brain theory and neural networks},
	volume={3361},
	year={1995},
	publisher={Cambridge, MA: MIT Press}
}

@inproceedings{bello2019attention,
	title={Attention augmented convolutional networks},
	author={Bello, Irwan and Zoph, Barret and Vaswani, Ashish and Shlens, Jonathon and Le, Quoc V},
	booktitle={Proceedings of the IEEE/CVF international conference on computer vision},
	pages={3286--3295},
	year={2019}
}

@inproceedings{vaswani2017attention,
  title     = {Attention is All You Need},
  author    = {Vaswani, Ashish and Shazeer, Noam and Parmar, Niki and Uszkoreit, Jakob and Jones, Llion and Gomez, Aidan N and Kaiser, {\L}ukasz and Polosukhin, Illia},
  booktitle = {Advances in Neural Information Processing Systems (NeurIPS)},
  volume    = {30},
  year      = {2017}
}

@inproceedings{dosovitskiy2021vit,
  title     = {An Image is Worth 16x16 Words: Transformers for Image Recognition at Scale},
  author    = {Dosovitskiy, Alexey and Beyer, Lucas and Kolesnikov, Alexander and Weissenborn, Dirk and Zhai, Xiaohua and Unterthiner, Thomas and Dehghani, Mostafa and Minderer, Matthias and Heigold, Georg and Gelly, Sylvain and Uszkoreit, Jakob and Houlsby, Neil},
  booktitle = {International Conference on Learning Representations (ICLR)},
  year      = {2021}
}

@inproceedings{liu2021swin,
  title     = {Swin Transformer: Hierarchical Vision Transformer using Shifted Windows},
  author    = {Liu, Ze and Lin, Yutong and Cao, Yue and Hu, Han and Wei, Yixuan and Zhang, Zheng and Lin, Stephen and Guo, Baining},
  booktitle = {Proceedings of the IEEE/CVF International Conference on Computer Vision (ICCV)},
  pages     = {10012--10022},
  year      = {2021}
}

@inproceedings{pan2022acmix,
	title={On the integration of self-attention and convolution},
	author={Pan, Xuran and Ge, Chunjiang and Lu, Rui and Song, Shiji and Chen, Guanfu and Huang, Zeyi and Huang, Gao},
	booktitle={Proceedings of the IEEE/CVF conference on computer vision and pattern recognition},
	pages={815--825},
	year={2022}
}

@article{li2023uniformer,
	title={Uniformer: Unifying convolution and self-attention for visual recognition},
	author={Li, Kunchang and Wang, Yali and Zhang, Junhao and Gao, Peng and Song, Guanglu and Liu, Yu and Li, Hongsheng and Qiao, Yu},
	journal={IEEE Transactions on Pattern Analysis and Machine Intelligence},
	volume={45},
	number={10},
	pages={12581--12600},
	year={2023},
	publisher={IEEE}
}

@inproceedings{wei2023dmformer,
	title={DMFormer: Closing the gap between CNN and vision transformers},
	author={Wei, Zimian and Pan, Hengyue and Li, Lujun and Lu, Menglong and Niu, Xin and Dong, Peijie and Li, Dongsheng},
	booktitle={ICASSP 2023-2023 IEEE International Conference on Acoustics, Speech and Signal Processing (ICASSP)},
	pages={1--5},
	year={2023},
	organization={IEEE}
}

@article{krizhevsky2012imagenet,
	title={Imagenet classification with deep convolutional neural networks},
	author={Krizhevsky, Alex and Sutskever, Ilya and Hinton, Geoffrey E},
	journal={Advances in neural information processing systems},
	volume={25},
	year={2012}
}

@article{simonyan2014vgg,
	title={Very deep convolutional networks for large-scale image recognition},
	author={Simonyan, Karen and Zisserman, Andrew},
	journal={arXiv preprint arXiv:1409.1556},
	year={2014}
}

@inproceedings{he2016resnet,
	title={Deep residual learning for image recognition},
	author={He, Kaiming and Zhang, Xiangyu and Ren, Shaoqing and Sun, Jian},
	booktitle={Proceedings of the IEEE conference on computer vision and pattern recognition},
	pages={770--778},
	year={2016}
}

@article{hu2020senet,
  title   = {Squeeze-and-Excitation Networks},
  author  = {Hu, Jie and Shen, Li and Albanie, Samuel and Sun, Gang and Wu, Enhua},
  journal = {IEEE Transactions on Pattern Analysis and Machine Intelligence},
  volume  = {42},
  number  = {8},
  pages   = {2011--2023},
  year    = {2020}
}

@inproceedings{wang2020ecanet,
  title     = {{ECA-Net}: Efficient Channel Attention for Deep Convolutional Neural Networks},
  author    = {Wang, Qilong and Wu, Banggu and Zhu, Pengfei and Li, Peihua and Zuo, Wangmeng and Hu, Qinghua},
  booktitle = {Proceedings of the IEEE/CVF Conference on Computer Vision and Pattern Recognition (CVPR)},
  year      = {2020}
}

@article{park2018bam,
	title={Bam: Bottleneck attention module},
	author={Park, Jongchan and Woo, Sanghyun and Lee, Joon-Young and Kweon, In So},
	journal={arXiv preprint arXiv:1807.06514},
	year={2018}
}

@inproceedings{woo2018cbam,
  title     = {CBAM: Convolutional Block Attention Module},
  author    = {Woo, Sanghyun and Park, Jongchan and Lee, Joon-Young and Kweon, In So},
  booktitle = {Proceedings of the European Conference on Computer Vision (ECCV)},
  pages     = {3--19},
  year      = {2018}
}

@misc{krizhevsky2009cifar,
  title       = {Learning Multiple Layers of Features from Tiny Images},
  author      = {Krizhevsky, Alex},
  year        = {2009},
  howpublished = {Technical Report, University of Toronto}
}

@inproceedings{paszke2019pytorch,
  title     = {{PyTorch}: An Imperative Style, High-Performance Deep Learning Library},
  author    = {Paszke, Adam and Gross, Sam and Massa, Francisco and Lerer, Adam and Bradbury, James and Chanan, Gregory and Killeen, Trevor and Lin, Zeming and Gimelshein, Natalia and Antiga, Luca and others},
  booktitle = {Advances in Neural Information Processing Systems (NeurIPS)},
  volume    = {32},
  year      = {2019}
}

@article{devlin2019bert,
  title   = {{BERT}: Pre-Training of Deep Bidirectional Transformers for Language Understanding},
  author  = {Devlin, Jacob and Chang, Ming-Wei and Lee, Kenton and Toutanova, Kristina},
  journal = {arXiv preprint arXiv:1810.04805},
  year    = {2019}
}

@misc{radford2018gpt,
  title       = {Improving Language Understanding by Generative Pre-Training},
  author      = {Radford, Alec and Narasimhan, Karthik and Salimans, Tim and Sutskever, Ilya},
  year        = {2018},
  howpublished = {OpenAI Technical Report}
}

@misc{radford2019gpt2,
  title       = {Language Models are Unsupervised Multitask Learners},
  author      = {Radford, Alec and Wu, Jeffrey and Child, Rewon and Luan, David and Amodei, Dario and Sutskever, Ilya},
  year        = {2019},
  howpublished = {OpenAI Technical Report}
}

@article{brown2020gpt3,
  title   = {Language Models are Few-Shot Learners},
  author  = {Brown, Tom B. and Mann, Benjamin and Ryder, Nick and Subbiah, Melanie and others},
  journal = {arXiv preprint arXiv:2005.14165},
  year    = {2020}
}

\end{document}